\documentclass[11pt]{article}
\usepackage{coling2020}
\usepackage{times}
\usepackage{url}
\usepackage{latexsym}
\usepackage{xcolor}
\usepackage{graphicx}
\usepackage{float}
\usepackage{hyperref}

\usepackage{amsmath}
\DeclareMathOperator{\softmax}{softmax}
\DeclareMathOperator{\LN}{LN}
\DeclareMathOperator{\MHAtt}{MHAtt}
\DeclareMathOperator{\FFN}{FFN}
\colingfinalcopy % Uncomment this line for the final submission

\title{Product Title Generation for Conversational Systems using BERT}

\author{Mansi Ranjit Mane, Shashank Kedia, Aditya Mantha, Stephen Guo, Kannan Achan  \\
  Walmart Labs, Sunnyvale, CA, USA \\
  {\{mansi.mane, shashank.kedia, aditya.mantha, sguo, kachan\}@walmartlabs.com}\\}

\date{}

\begin{document}
\maketitle
\begin{abstract}
  Through recent advancements in speech technology and introduction of smart devices, such as Amazon Alexa and Google Home, increasing numbers of users are interacting with applications through voice. E-commerce companies typically display short product titles on their webpages, either human-curated or algorithmically generated, when brevity is required, but these titles are dissimilar from natural spoken language. For example, ``Lucky Charms Gluten Free Break-fast Cereal, 20.5 oz a box Lucky Charms Gluten Free" is acceptable to display on a webpage, but ``a 20.5 ounce box of lucky charms gluten free cereal" is easier to comprehend over a conversational system. As compared to display devices, where images and detailed product information can be presented to users, short titles for products are necessary when interfacing with voice assistants. We propose a sequence-to-sequence approach using BERT to generate short, natural, spoken language titles from input web titles. Our extensive experiments on a real-world industry dataset and human evaluation of model outputs, demonstrate that BERT summarization outperforms comparable baseline models.
\end{abstract}

% \blfootnote{

%     \href{https://github.com/nlpyang/PreSumm}{https://github.com/nlpyang/PreSumm} 
%     \hspace{-0.65cm}  % space normally used by the marker
% }

% \blfootnote{
%     %
%     % for review submission
%     %
%     \hspace{-0.65cm}  % space normally used by the marker
%     Code/:\href{https://github.com/nlpyang/PreSumm}{Code} 

% }

\section{Introduction}
Smartphones and voice-activated smart speakers, such as Amazon Alexa, Google Home, and Apple Siri, have led to increased adoption of voice-enabled shopping experiences. In such voice-enabled shopping experiences, reducing user friction and saving time is key, especially for low consideration purchases and repeat grocery purchases. Display-based experiences typically utilize a short product title when presenting a product, but these short titles do not naturally fit in a typical conversational flow. For example, for a display-based experience showing “Jergens Natural Glow Daily Moisturizer, Medium to Tan, 7.5 oz” as a short title of the product may be okay but it is not suitable for voice-based applications as it is not a naturally spoken title. At the same time, display-based experiences have the added benefit of being able to show a lot of additional meta-data which is not possible in conversational systems. The product title for a conversational system needs to encapsulate the important information in a succinct, grammatically correct natural language sentence, such that it naturally fits with the conversational flow during the dialogue rounds between the user and the conversational system. E-commerce companies can have millions to billions of products in their ever-changing product catalogs, so it is typically prohibitively expensive to manually annotate naturally spoken titles for all products. In such industry settings, it is ideal to have a model that can generate naturally spoken titles for an evolving catalog. The primary goal of this work is to examine the methods and challenges to convert short titles of products into a naturally spoken language that is grammatically correct.

%   Issue of duplicate words is very common in machine translation and text summarization, this can cause negative customer experience. The articles 

This problem is similar to the \textit{text summarization} task which is well studied in natural language processing (NLP). However, using it for e-commerce application is not a trivial task. Text summarization approaches can be classified into two broad sub-categories, \textit{extractive} text summarization and \textit{abstractive} text summarization. Extractive text summarization-based approaches usually try to extract a few sentences (or keywords) from lengthy documents \cite{dorr2003hedge,Neto,NallapatiExtractive}. To generate natural language titles from web titles, model should have capability to generate conjunctions, articles etc. at the appropriate position.

%Traditional methods have used a decision on which text to delete from the original text to generate summaries \cite{dorr2003hedge} \cite{Neto}. Since synonyms and word understanding of models were not very well developed, most methods were of an extractive nature, i.e., they relied on extracting information from original text rather than paraphrasing text.
% Initial methods attempted to extract important information in the input sentence and use it directly in summaries.

Abstractive text summarization attempts to understand the content of the document and produce summaries which may contain novel words or phrases. Recurrent Neural Networks(RNN) ~\cite{LSTM,GRU} based sequence to sequence (seq2seq) models \cite{sutskever2014sequence} or recently developed attention models ~\cite{vaswani2017attention} have been shown to perform well on abstractive summarization tasks ~\cite{NallapatiAbstractiveCNN}. However these models tend to generate repetitive words and this can cause negative customer experience in voice e-commerce applications. For traditional text summarization tasks we can introduce novel words which are not part of the input sequence but in an e-commerce setting paraphrasing of factual details like brand, quantity, etc. may imply a different product.
In addition to this, new products are added to e-commerce product catalogs continuously, which can introduce out-of-vocabulary words. The summarization model should be able to generalize to these words.

% While it's okay to have words replaced by there synonyms in normal text summarization, for product title generation, proper nouns such as brand and quantity need to be kept unchanged in generated titles.  
% With the improvement in computational feasibility of Recurrent Neural Networks (RNN) \cite{LSTM} \cite{GRU}, text summarization using RNNs to capture the context of the text, especially by using Sequence to sequence(Seq2seq) models\cite{sutskever2014sequence}, exploded and lead to increased research in text summarization \cite{NallapatiExtractive} \cite{DBLP:journals/corr/RushCW15}. In seq2seq methods, an input text is sent through what is called an encoder which learns to capture the important aspects or context of the input text. This context is then used to train a decoder that outputs a sequence that should form the relevant output sequence for the input model. Attention mechanisms  \cite{bahdanau2014neural}, improved the seq2seq models further\cite{vaswani2017attention} by providing extra context around which input words to focus on at each decoding step.

In this paper, we investigate application of text summarization techniques to voice based e-commerce application. Our major contributions are
\begin{enumerate}
\item We adapt different state-of-the-art NLP models to a real-world e-commerce dataset with limited labels.
\item We perform extensive evaluation of these models on established evaluation metrics, as well metrics relevant to our application. We also perform human judgement evaluation.
\end{enumerate}
% 1) We adapt different state-of-the-art NLP models to a real world e-commerce dataset with limited labels. 2) We perform extensive evaluation of these models in terms of established evaluation metrics as well metrics relevant to e-commerce applications and human evaluation. 
 
 The following sections provides a summary of related work followed by a description of methods applied to convert web-based short titles of products (sequence of words in English) into more naturally spoken summary titles (sequence of words in English) for voice-based applications. In our problem setting, we are more interested in building an abstractive text summarization based model that can generate novel words in the decoded summary.  Section~\ref{sec:experiment} provides the salient features of the dataset and implementation details of the methods described earlier. This is followed by a discussion of results observed and conclusion.

%  Some key challenges like factoring in repetitive duplicate words and the model can generalize to out of vocabulary words in the input sequence is also discussed.
% \begin{productize}
%     \item Why do we need this - problem statement
%     \item Abstractive vs Extractive summarization - inserting new words 
%     \item Challenges: Generalizing to new words in the titlehttps://www.overleaf.com/project/5f129d520c36e30001052e0d
%     \item Challenges:Grammar
%     \item Challenges:Duplicate words
%     \item Challenges: Penalization
% \end{itemize}

\section{Related Work}

Text summarization is a long-studied problem in natural language processing. With the advent of deep learning based approaches, seq2seq models have proven highly successful in abstractive text summarization. Some of these models and relevant developments in the field are mentioned below.

Ptr-Net~\cite{pointernet} develops on the seq2seq model with attention for the summarization task. It uses the concept of the pointer network introduced in Vinyals et al.~\shortcite{vinyals2015pointer} to decide which words from the main text should directly be copied to the summary. This helps to preserve the important factual information from the input text and also assists in handling out-of-vocabulary words. Ptr-Net model also adds coverage loss, which examines the difference between the attentions of previous words generated and the current attention, in an attempt to fix the issue of word repetition, a persistent issue in seq2seq models. Gehrmann et al.~\shortcite{gehrmann2018bottom} try to improve the fluency of the generated text through various constraints applied during model training. Soft constraints on the size of text are used to constrain the length of generated descriptions, while constraints on the output probability distribution of words ameliorates word repetition.

Development in language models have subsequently led to increased use of pre-trained models such as BERT~\cite{bert} and GPT-2~\cite{radford2019language}, which are trained on huge text corpuses and are used to generate the word embeddings for input texts. While Khandelwal et al.~\shortcite{khandelwal2019sample} examines the feasibility of pre-trained language models in a low-data setting and moves away from a seq2seq framework, the work of Liu and Lapata~\shortcite{presum} use BERT in a seq2seq model to summarize data. Details of this model are discussed further in Section\ref{sec:method}. In our study, this is the primary model adapted for our use-case.

Text summarizarion finds natural application in e-commerce where products may have a long description but only the salient features of a particular product is what the end user is in interested in. Increased interaction with mobile devices and voice based interfaces such as Amazon Alexa and Google Home present new challenges as the product title now need to be succinct. There has been development in rule based methods as in Camargo de Souza et al.~\shortcite{camargo-de-souza-etal-2018-generating}. Deep learning based methods find natural application and attempt to use multi-model information (images and text) to generate product title. Chen et al.~\shortcite{chen2019towards} generates personalized product titles utilizing user personas and an external knowledge base. Mathur et al.~\shortcite{mathur-etal-2018-multi} attempts at generating titles in different languages for the same product. Sun et al.~\shortcite{sun2018multi} develops further on the work of Ptr-Net using a separate encoder network for important attributes like quantity, brand of products and then then uses 2 pointers to decide where to copy data from. The work of Zhang et al.~\shortcite{zhang2019multi} attempts a novel method of generating short descriptions for e-commerce use case and uses multimodal information. A word sequence symbolizing description and the image of a product in the catalog are chosen as descriptors for the product and are used to generate short titles for the product. An adversarial network based approach is then used to decide if the generated title is machine or human generated to improve the quality of generated titles and make it more human like.
% \section{Problem Formulation}\label{sec:Problem}

% In this section, we formulate the problem of automatic natural language title generation. The goal of this task is to build a system that can automatically generate natural language titles of products that are easy to interpret in a voice-enabled shopping experience. Given the short web-title represented as a sequence of words $w = \{w_1, w_2, .... w_n\}$, the goal of this system is to generate the natural language title $y = \{y_1, y_2, .... y_m\}$ that is succinct of the short web-title.
\section{Methods}\label{sec:method}
In this section, we formulate the problem of automatic natural language title generation and discuss various approaches that can be used to solve this problem.

\textbf{Problem Definition:}
The goal of this task is to build a system that can automatically generate natural language product titles which are easily interpretable in a voice-enabled shopping experience. Given the short web-title $w$ represented as a sequence of words $w = \{w_1, w_2, .... w_n\}$, the goal of this system is to generate the corresponding natural language title $y = \{y_1, y_2, .... y_m\}$. 

In the following sub-sections, we discuss various models which we apply for the automatic natural language title generation task.
% In this section we discuss various approaches that can be used for the automatic natural language title generation task:
\subsection{seq2seq + Attention}
Consider a sequence of input tokens $w_i$ fed into an encoder (LSTM) producing a sequence of encoder hidden states $h_i$. The decoder receives the word embeddings of the previous words and has a decoder state $s_t$ at time step $t$. The attention distribution\cite{bahdanau2014neural} $a^t$ is computed as shown in following equations:
\begin{equation}\label{eq:attention1}
    e_i^t = v^t.\tanh(W_h h_i + W_t.s_t + b_{attn})
\end{equation}
\begin{equation}\label{eq:attention2}
    a^t = \softmax(e^t)
\end{equation}
where $v$, $W_h$, $W_s$, and $b_{attn}$ are learnable parameters. The attention distribution is used to produce a weighted sum of the encoder hidden states, known as the context vector $h_t^*$ computed by: 
\begin{equation}
  h_t^* = \sum_{i} a_i^t h_i  \\
\end{equation}

The context vector $h_t^*$ and decoder state $s_t$ is fed through linear layers to get the vocabulary distribution $P_{vocab}(w)$. The network is trained end-to-end using the negative log-likelihood of the target word $w_t^*$ at each timestep.
\begin{equation}
  loss_{t}= -\log P_{vocab}(w_t^*)
\end{equation}
\subsection{Ptr-Net}
\textit{Ptr-Net}~\cite{pointernet} is a hybrid between the baseline seq2seq with attention and a pointer network. In the this model, the generation probability $p_{gen}$(or the probability of using a new word in the output) depends on context vector $h_t^*$ and attention distribution $a^t$, and is computed as follows:
\begin{equation}
  p_{gen}= \sigma{(w_{h^{*}}^{T}h_t^* + w_s^T s_t + w_x^T x_t + b_{ptr})}
\end{equation}
where $w_{h^{*}}$, $w_s$, $w_x$, and $b_{ptr}$ are learnable parameters. $p_{gen}$ is used to choose between generating a word from the vocabulary by sampling from $P_{vocab}$, or copying a word from the input sequence by sampling from the attention distribution $a_{t}$. The final probability distribution over the vocabulary is computed by:
\begin{equation}\label{eq:nll}
  P(w)= p_{gen}. P_{vocab}(w) + (1 - p_{gen}) \sum_{i:w_i=w} a_i^t 
\end{equation}
The model is trained end-to-end similar to \textit{seq2seq + attention} using the negative log-likelihood of the target word $w_t^*$ as the loss function.
\subsection{Transformer}
% For every input word, query, Key and Value vectors are created.
Transformers~\cite{vaswani2017attention} are attention-based models, where the relationship between a given word $w$ and the context is modeled through multi-head attention. Each layer in a transformer consists of multi-head attention ($\MHAtt$) followed by a layer norm {$\LN$} and feed forward network $\FFN$ as shown in Equation~\ref{eq:attn}.

\begin{equation}\label{eq:attn}
    \tilde{h}^l = \LN(h^{l-1} + \MHAtt(h^{l-1})
\end{equation}
\begin{equation}
     h^l = \LN(\tilde{h}^l + \FFN(\tilde{h}^l)
    % h = LN(h^{l} + MHAtt(h^{l}))
\end{equation}

The final layer representation from the encoder is given to the decoder, and the decoder is trained using negative log likelihood with the target word $w_t^*$.

\subsection{BERT}
\begin{figure}[H]
\centering
\includegraphics[height= 4in,width=4.7in]{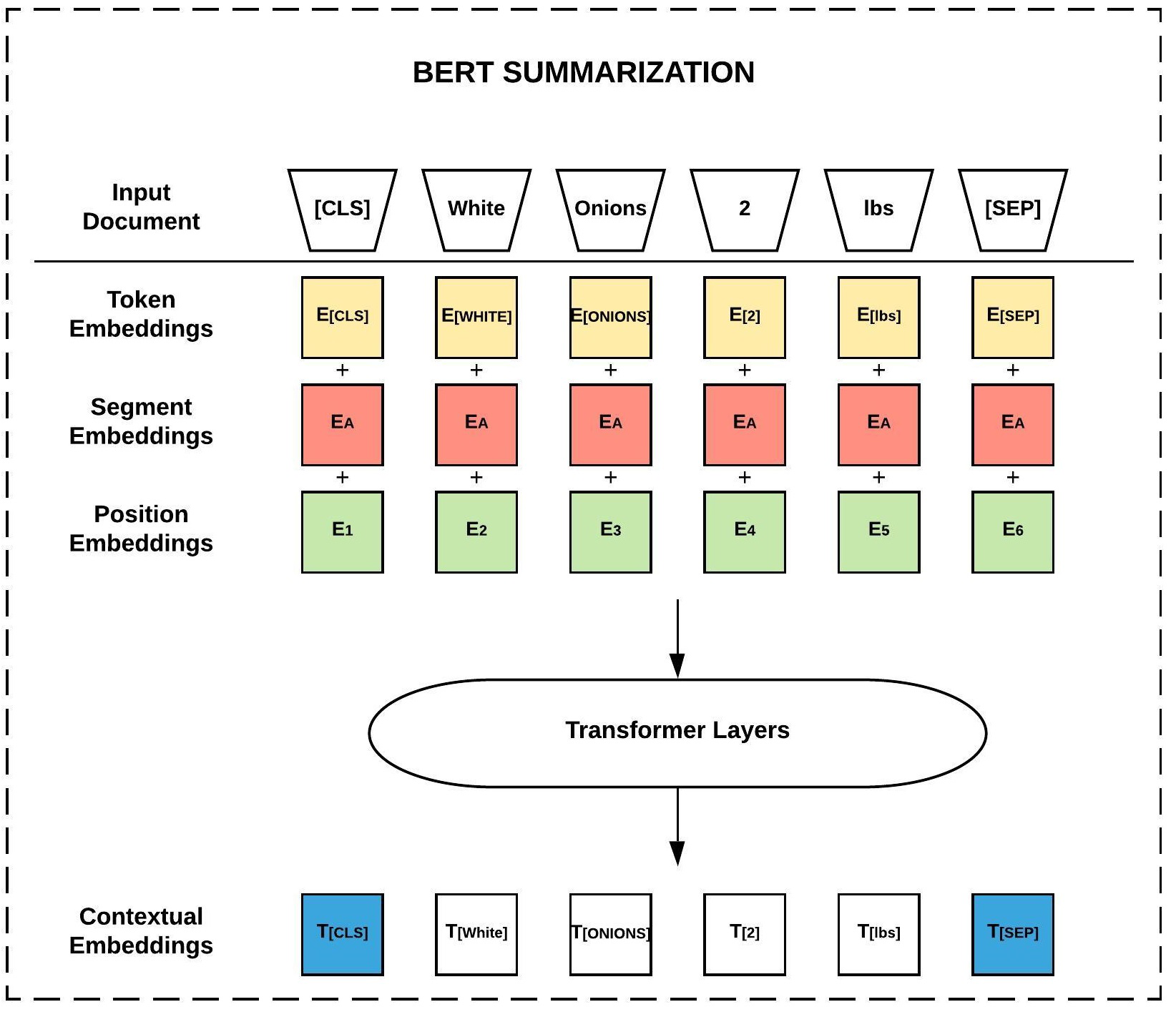}
\caption{Architecture for the BERT summarization model. Input document at the top is a sequence of words from the web title.}
\label{fig:bert}
\end{figure}

We use BERT (Bidirectional Encoder Representations from Transformer)~\cite{bert} to encode the web titles. BERT is trained on a large text corpora using unsupervised tasks ( masked language modelling and next sentence prediction). BERT uses token, segment, and position embeddings to represent input tokens. Segment embeddings are used in pairwise tasks to differentiate between segments (e.g., question and answer in SQUAD tasks). This input representation is then fed into multiple transformer layers.

Liu and Lapata~\shortcite{presum}, modify the BERT formulation for the text summarization task. A [CLS] token is used at the start of a sentence, and the representation of this token is used as the sentence representation.  $E_A$ and $E_B$ are used as segment embeddings for odd and even sentences, respectively.  Position embeddings for sentences larger than 500 words are learnt as model parameters. We adapt to this format to represent data. For our use case, web product titles are one sentence long, hence the segment embedding $E_B$ is not used.

We insert [CLS] and [SEP] tokens in input web title $w = [w_1, w_2 ...w_n]$ as shown in Figure~\ref{fig:bert}.The input representation  $[x_1, x_2 ... x_n]$ for transformer is then prepared by adding position ($E_{pos}$), token ($E_{token}$), and segment embedding ($E_{A}$) for corresponding words

\begin{equation}
    x_i = E_{token} + E_{pos} + E_{A}
\end{equation}

 Encoder then transforms this input representation using transformer layers which applies transformation as shown in Equation ~\ref{eq:attn} on input representation $x$. In Equation~\ref{eq:attn}, $h^{0}$ represents input representation $x_i$. The continuous representation from the final layer of BERT $[z_1,z_2,..,z_n]$ is then given to the decoder. While pretrained BERT is used as the encoder, an 8-layer transformer, randomly initialized, is used as the decoder. We train the decoder to generate summaries using true labels in ground truth data using the framework for abstractive summarization, as in ~\cite{pointernet}, without using the coverage and copy mechanism. We use the Adam optimizer with the following learning rate schedule for the encoder and decoder ~\cite{presum}: \\ 
\begin{equation}
    lr_{e} = \tilde{lr}_{e}.\min(step^{-0.5}, step.warmup_e^{-1.5}) 
\end{equation}
\begin{equation}
    lr_{d} = \tilde{lr}_{d}.\min(step^{-0.5}, step.warmup_d^{-1.5})
\end{equation}

We set $lr_{e} = 2e^{-3}$, $lr_{d} = 0.2$, $warmup_e = 20,000$, and $warmup_e = 10,000 $. 

\section{Experimental Setup}\label{sec:experiment}
\subsection{Dataset}
 We use a proprietary dataset from one of the largest E-commerce retailers in the world. Our dataset consists of only 19,269 pairs of web product titles along with their corresponding voice titles. Details of the dataset have been provided in Table~\ref{table:dataset}. The web product titles are either entered by merchants or algorithmically generated for certain categories while publishing products on e-commerce website. The voice titles for the corresponding web titles are manually created by human annotators through a crowdsourcing platform. Some examples of web titles and corresponding voice titles are shown in Table~\ref{table:examples}.
 
\begin{table}[H]
\centering
\begin{tabular}{ |p{0.6\linewidth}|p{0.4\linewidth}|}
\hline
\textbf{Web titles} (Input Sequence of Words) & \textbf{Voice titles} (Output Sequence of Words)\\
\hline
El Monterey Beef \& Cheese Burritos 8 ct bag a family size pack El Monterey Beef and cheese & a family size pack of 8 El Monterey Frozen Beef And Cheese Burrito \\
\hline
Paas Magical Color Cup Egg Decorating Kit a pack Paas Magical Color Cup & a pack of Paas Magical Color Cup Egg Decorating Kit \\
\hline
Wonderful Roasted \& Salted Pistachios 8 oz. Bag a bag Wonderful Roasted and salted & an 8 ounce bag of Wonderful Roasted And Salted Pistachios  \\\hline
\end{tabular}
\caption{Examples of input web titles along (left) and desired output voice titles (right).}\label{table:examples}
\end{table}

There are certain key differences in the characteristics of web titles and voice titles. Important distinctions are listed below:
\begin{itemize}
    \item Web titles often contain abbreviations for units of measurement for succinctness, e.g., Row 3 in Table~\ref{table:examples} mentions "8 oz. bag". However, the voice title should contain the corresponding natural language word "ounce".
    \item Web titles may or many not contain articles, but voice titles need to have grammatically correct articles, conjunctions, etc. For example, refer to Row 1 in Table~\ref{table:examples}.
    \item Web titles sometimes contain specific product attributes, such as brand or quantity. These product attributes may have altered positions in the voice title, but the attribute phrase needs to be retained exactly in its entirety. 
    \item As shown in Table~\ref{tab:dataset}, the average voice title length is $11.39$ tokens. Voice titles need to be short and succinct, as this information is spoken through a voice device to the end-user.
\end{itemize}

\begin{table}[H]
\centering
\begin{tabular}{|c|c|}
\hline
avg. web title length & 15.3352\\
avg. voice title length  & 11.3886\\
avg. \# unique words in web title  &  13.0805\\
avg. \# unique words in voice titles  & 11.3009\\
avg. \# of novel one grams in voice title & 4.0138\\
\# train examples & 13874\\
\# val examples  & 1926\\
\# test examples & 3469\\\hline
\end{tabular}
\caption{Dataset Statistics.}\label{tab:dataset}
\label{table:dataset}
\end{table}

% The web data contains short forms for units of measurement, such as "oz" instead of "ounce" and "ct" instead of "count". These translations are important in a voice setting as we need to communicate the corresponding full word related to units. We also need the articles used in generated voice title grammatically correct. Another important aspect is to maintain the brand information, which needs to be provided at the time of user communication.

We observed that web titles do not have specific details like brand for few products, hence we append additional product metadata when available to web title. This metadata contains attributes such as brand, container type, and size (whenever available).

The dataset is randomly partitioned, into 13874 train examples, 1926 validation examples, and 3469 test examples.

% The main different between our dataset and other summarization based datasets is  use of quantities. Since customer may purchase the product without looking at visual display the quantity should be retained in translation.

\subsection{Implementation Details}
We use the pytorch `bert-base-uncased' version of BERT for the encoder along with the subword tokenizer\footnote{ \url{https://github.com/nlpyang/PreSumm/}}.  In the decoder, the transformer has 768 hidden units and 8 layers, while the feed-forward layers have size 2,048. The learning rate used is as mentioned in Section~\ref{sec:method}, with a batch size of 256, and the model was trained for 35,000 steps. We used beam search with size $5$ and $\alpha = 0.95$ for decoding. Decoding is done until end of sequence token is emitted.  we also block repeat trigrams ~\cite{presum}. We use a minimum length of 4 for decoding and a maximum length of 50 for decoding. A checkpoint model was saved every 2,000 steps, with the best performing checkpoint model on validation data being used to report performance on the test data.

We compare the BERT model with seq2seq, Ptr-Net, Ptr-Net + Coverage, and the Transformer model as baselines. For the implementation of seq2seq, Ptr-Net, and Ptr-Net + Coverage, the implementation of~\cite{pointernet} was used to generate the results\footnote{ \url{https://github.com/abisee/pointer-generator}}. 

The implementation details of the various baselines is provided below:
\begin{itemize}

    \item \textbf{seq2seq}: Stanford core nlp PTBtokenizer is used to tokenize the data, which is converted into story format as in the popular CNN-Daily mail dataset \cite{NallapatiAbstractiveCNN} for text summarization. We use default parameters from authors and change minimum length in decoder to 50 and maximum length in decoder to 35, as those are the corresponding maximum lengths of augmented web title and voice title respectively. When decoding test data, a beam search of beam size 4 is used to generate the predicted title. A minimum length of 5 is set for the prediction title. 
    
    \item \textbf{Ptr-Net}: The Pointer Net uses the same parameters as seq2seq model, and the validation set is used to identify the optimal training checkpoint for the model.
    
    \item \textbf{Ptr-Net + Coverage}: We use the default implementation of the authors and train the model in a 2-step training process. First, the pointer net model is trained without any coverage loss. Using validation loss, the best model is extracted. We then add the coverage loss term, and train the model again using the previously mentioned best model as warm-up. 
    
    \item \textbf{Transformer}: We use a 6-layer transformer encoder with 512 hidden size and 2048 dimensional feed-forward layer. For the decoder, we use the same configuration as BERT.  The learning rate and other hyper-parameters are obtained from ~\cite{presum}.

\end{itemize}

\subsection{Evaluation Metrics}
We use the following metrics to evaluate the above proposed model and baselines:
\begin{itemize}
    \item \textbf{ROUGE} (Recall-Oriented Understudy for Gisting Evaluation): measures overlap between the candidate summary and ground truth using precision, recall, and F-1 scores. However, ROUGE does not give a clear idea about repetitions or duplicates in the generated summary. We report F-1 ROUGE score at 1, 2, and L(Longest Common Subsequence). \cite{lin2004rouge}
    \begin{itemize}
    \item \textbf{ROUGE-1} refers to the overlap of the unigrams
    \item \textbf{ROUGE-2} refers to the overlap of the bigrams
    \item \textbf{ROUGE-L} takes into account sentence level structure similarity naturally and identifies the longest co-occurring in-sequence n-grams automatically.
    \end{itemize}
    \item\textbf{Avg. \# duplicate 1-grams} - Number of duplicate one-grams between the ground truth summary and candidate summary.
    \item\textbf{Human Evaluation} - 3 human annotators were provided 100 random samples of model-output titles and were asked to rate the title on a scale of 1-5, based on product relevance(i.e., if the product is the same), grammatical correctness, and correct preservation of important attributes (such as brand name, quantity, and unit of measurement). The average score of the annotators is taken as the judgement score for the model. This evaluation was performed for only the Transformer and BERT Abstractive model outputs.
\end{itemize}

\section{Results and Observations}

\begin{table}
\centering
\begin{tabular}{|c|c|c|c|c|c|}
\hline
\textbf{Method} & \textbf{R-1} &  \textbf{R-2} &  \textbf{R-L}&  \textbf{avg. \# of duplicates} & \textbf{Human Evaluation}\\\hline
seq2seq + attention & 0.7951  & 0.6607 & 0.7883 & 0.257135 & X\\
Ptr-Net & 0.8965  & 0.8053  & 0.8956 & 0.239839 & X \\
Ptr-Net with Coverage & 0.8917  & 0.8042  & 0.8800 & 0.3041 & X\\
Transformer  & \textbf{0.92715} & \textbf{0.85812}  &\textbf{0.92012} & 0.187950 & 4.19 \\
Bert Abstractive  & 0.92298 & 0.84092 & 0.91383 & \textbf{0.17262} & \textbf{4.35} \\ \hline
\end{tabular}
\caption{Evaluation Results. R-1, R-2, and R-L denote ROUGE-1, ROUGE-2, and ROUGE-L metrics.}\label{tab:results}
\end{table}

Table~\ref{tab:results} provides a summary of model performance on different evaluation metrics. We observe that transformer and BERT have ROUGE-1 F1 scores of $0.92715$ and $0.92298$ respectively, outperform seq2seq-based approaches in terms of both ROUGE and avg. \# duplicates metrics. The Ptr-Net model outperforms basic seq2seq+attention approach, however, paradoxically Ptr-Net with coverage underperforms compared to Ptr-Net model. The coverage loss is present to specifically address the issue of repetition of words and instead of fixing it, lead to an increase in the avg.\# of duplicates to $0.3041$ from $0.239839$. While transformer model does have a better ROUGE score than BERT model, BERT has lower repeated words in output ($0.17262$ compared to $0.18795$) which has a greater impact on the title qualitatively. Since, BERT is pre-trained on a large corpus, it should be able to generalize better especially in low data scenarios. Human evaluation of generated titles reinforce this, showing that BERT performs better than transformer based model on output quality. 

%While pointer generator based approaches perform decent with respect to ROUGE based metrics, Ptr-Net model has some issues in picking the right article for the sentence like “a” vs “an” while BERT based approach seems to get such detailed nuances better as it is pre-trained on a huge corpus of textual data. We also observed that BERT performs slightly better when compared to pointer generator network in not showing redundant duplicate words. 

\begin{table}[H]
\centering
\begin{tabular}{ |p{0.025\linewidth}|p{0.3\linewidth}| p{0.28\linewidth}|p{0.28\linewidth}|}
\hline
& \textbf{Web title} & \textbf{Ground truth} & \textbf{Model prediction}\\
\hline
1 & White Onions2 lbs a bag  premium & 
a 2 pound bag of white onions& 
a 2 pound bag of white onions \\
\hline
2 & Great Value Large Grade AA, 6 Eggs a carton Great Value Large Grade AA & a half dozen carton of Great Value Large Grade AA Eggs & a 6 count carton of great value large grade aa eggs \\
\hline
3 & Lucky Charms Gluten Free Breakfast Cereal, \textcolor{blue}{\textbf{20.5 oz}} a box Lucky Charms Gluten Free & 
a \textcolor{red}{\textbf{19.3 ounce}} box of Lucky Charms Gluten Free Cereal &
a \textcolor{blue}{\textbf{20.5 ounce}} box of lucky charms gluten free cereal \\
\hline
4 & Hostess Donettes \textcolor{blue}{\textbf{Frosted Mini Donuts}}, 6 ct, 3 oz a pack hostess donettes &
a pack of 6 hostess donettes \textcolor{red}{\textbf{mini donuts}} & 
a pack of 6 hostess donettes \textcolor{blue}{\textbf{frosted mini donuts}} \\ \hline
5 & Pork Butt Steaks Large, Tray, \textcolor{blue}{\textbf{3.1 - 5.1 lbs a tray}} &
a \textcolor{red}{\textbf{34.4 ounce tray}} of pork butt & 
a \textcolor{blue}{\textbf{3.1 to 5.1 pound}} tray of pork butt steaks \\ \hline
6 & Pork Cube Steaks, Tray, \textcolor{blue}{\textbf{0.45 - 1.35 lbs}} a tray &
a \textcolor{red}{\textbf{12 ounce tray}} of ground pork &
a \textcolor{blue}{\textbf{1.35 pound}} tray of pork cubes\\ \hline
\end{tabular}
\caption{BERT Summariazation - Good Model Predictions.}\label{table:success}
\end{table}

\begin{table}[H]
\centering
\begin{tabular}{ |p{0.025\linewidth}|p{0.3\linewidth}| p{0.28\linewidth}|p{0.28\linewidth}|}
\hline
 & \textbf{Web Title} & \textbf{Ground truth} & \textbf{Model prediction} \\
\hline
1 & Yoo-Hoo Chocolate Milk Fridge Pack, 12 pk a pack Yoo Hoo \textcolor{blue}{\textbf{Chocolate Fridge Pack}} & a 12 count pack of yoo hoo \textcolor{blue}{\textbf{chocolate milk fridge pack}} & a pack of 12 yoo hoo \textcolor{red}{\textbf{chocolate bar}} \\
\hline
2 & Produce Unbranded Eat Your Vegetables Blend 7 Oz a pack &
a 7 ounce pack of veggie blend snack &
a 7 ounce pack of \textcolor{red}{\textbf{produce produce produce}} baby food blend
  \\
\hline
3 & Diet 7UP, \textcolor{blue}{\textbf{0.5 L}}, 6 pack a pack  & a 6 pack of \textcolor{blue}{\textbf{.5 liter diet}} 7up & a 6 pack of \textcolor{red}{\textbf{0.5 fluid ounce}} diet 7 up \\
\hline
4 & Garlic, each (1 bulb) a garlic & garlic sold individually & a \textcolor{red}{\textbf{pound}} of garlic sold individually \\\hline
5 & Harvestland Chicken Breast, \textcolor{blue}{\textbf{1.5-2 lbs.}} a tray perdue harvestl and free range & a \textcolor{blue}{\textbf{1.5 to 2 pound}} tray of Perdue Harvestland Free Range Chicken Breasts & a \textcolor{red}{\textbf{1.5 to 1.5 pound}} tray of perdue harvestland free range chicken breast \\
\hline
\end{tabular}
\caption{BERT Summariazation - Bad Model Predictions.}\label{table:failure}
\end{table}

Table~\ref{table:success} and Table~\ref{table:failure} lists examples of BERT Abstractive Model prediction on the test dataset where the model performs well and poorly, respectively. The corresponding ground truth and web titles have also been provided for comparison. The model seems to have repeated words in certain cases, for example Row 2 of Table~\ref{table:failure}.Given that most of the data has been trained on products with ounce and pound as the units of measurement, it can be seen that liter is incorrectly converted to fluid ounce by the model in Row 3 of Table~\ref{table:failure} and pound is added to a single bulb of garlic in Row 4 of Table~\ref{table:failure}. However, it can be observed that in some cases, the model clearly does better than ground truth evaluation and even fixes the incorrect quantities in ground truth (rows 3-6 in Table~\ref{table:success}). The model is able to add important attributes like \textit{frosted} in the case of row 4 of Table~\ref{table:success}. From row 2 of Table~\ref{table:success} we can see that the model is able to maintain the brand name as it is like \textit{great value} and provide the correct measurements units for the different products like \textit{6 count} or \textit{pack of 6}. Thus the model is able to fulfill requirements of preserving quantity and brand between web and voice titles.

We observe that overall BERT based model performs better \textit{both} quantitatively and qualitatively in maintaining factual details in the output title and also reducing repeated words in the output.

% Padding lenth change experiment details
% Warm up schedule details
% Freezing and non freezing word embedding layers
% Aditya

% \begin{table}
% \centering
% \begin{tabular}{c|c|c|c|c}
% \hline
% \textbf{Method} & \textbf{R-1} &  \textbf{R-2} &  \textbf{R-L} &  \textbf{avg \# duplicates}\\
% \hline
% seq2seq + attention & 0.8206 & 0.6997 & 0.8138 & 0.20899\\
% Pointer Net & 0.8961 & 0.8123 & 0.8865 & 0.2876 \\
% Pointer Net with coverage & 0.8961 & 0.8123 & 0.8865 & 0.2876\\
% Transformer  & x & x & x & x\\
% Bert Abstractive & 0.83114 & 0.66970 & 0.81773 & x\\ \hline
% \label{table:dataset}
% \end{tabular}
% \caption{Results: Input sequence is web titles without any additional attributes }\label{tab:results}

% \end{table}

% Observations in results- Mansi/Aditya/sahasnk 
% duplicates 
% articles 
% grammar

\section{Conclusion}
In this paper, we studied the problem of generating succinct, grammatically correct voice titles for products in a large e-commerce catalog with limited labels. We evaluate 4 different baselines and demonstrate that BERT summarization can generate good titles through ROUGE metrics and human evaluation, even when there is extremely limited data. Generating personalized titles for different user segments based on rich user metadata and incorporating web data with additional product attributes that may be product dependent are some directions to extend this work.

\bibliographystyle{coling}
\bibliography{coling2020}

\end{document}